\documentclass[10pt,a4paper]{article} 

\usepackage{layout}
\usepackage{ifthen}
\usepackage{url} 
\usepackage{doi}

\makeindex

\usepackage{amsmath}
\usepackage{amsfonts}
\usepackage{subfigure}
\usepackage{graphicx}

\newcommand\sectionname{Section}
\newcommand\equationname{Eq.}
\newcommand\eg{\textit{e.g.},}
\newcommand\ie{\textit{i.e.}}
\newcommand\cf{\textit{cf}}
\newcommand\emphterm[1]{\emph{#1}}
\newcommand\numberOfExamples{n}
\newcommand\dimension{d}
\newcommand\landau{\mathcal{O}}
\newcommand\hypothesis{g}
\newcommand\hypothesisSpace{\mathcal{H}}
\newcommand\categoryName[1]{\textit{#1}}
\newcommand\newterm[1]{\emph{#1}\index{#1}}
\newcommand\newtermex[2]{\emph{#1}\index{#2}}
\newcommand\newtermabb[2]{\emph{#1 (#2)}\index{#1}\index{#2|see{#1}}}

\newcommand{\IncludeGraphicWithLongCaption}[5]{
	\begin{figure}[tb]
	\centering
	\includegraphics[width=#5\linewidth]{figures/#4}
	\caption[#2]{#3}
	\label{#1}
	\end{figure}
}

\usepackage{natbib}

\author{Erik Rodner\\University of Jena, Germany\\\url{firstname.lastname@uni-jena.de}}
\date{August 2011}
\title{Visual Transfer Learning: Informal Introduction and Literature Overview}

\begin{document}

\maketitle
\begin{abstract}
	Transfer learning techniques are important to handle small training sets and to allow for quick generalization even from only a few examples. The following paper is the introduction as well as the literature overview part of my thesis related to the topic of transfer learning for visual recognition problems.
\end{abstract}

\section{Motivation}
\label{sec:motivation}

	As humans we are able to visually recognize and name a large variety of object categories.
	A rough estimation of \citet{Biederman87:Rto} suggests that we know approximately $30.000$
	different visual categories, which corresponds to learning five categories per day, on average, in our
	childhood. Moreover, we are able to learn the appearance of a new category using few
	visual examples~\citep{Parikh10:RoF}.
	Despite the impressive success of current machine vision systems~\citep{Everingham10:PVO},
	the performance is still far from being comparable to human generalization abilities.
	Current machine learning methods, especially when applied to visual recognition problems,
	often need several hundreds or thousands of training examples to build an appropriate \emphterm{classification model}
	or \emphterm{classifier}.
	Transfer learning techniques try to reduce this still existing gap between human and machine vision. 

\begin{figure}[tb]
		\subfigure[]{
			\label{fig:zipfbox}
			\includegraphics[angle=270,width=0.485\linewidth]{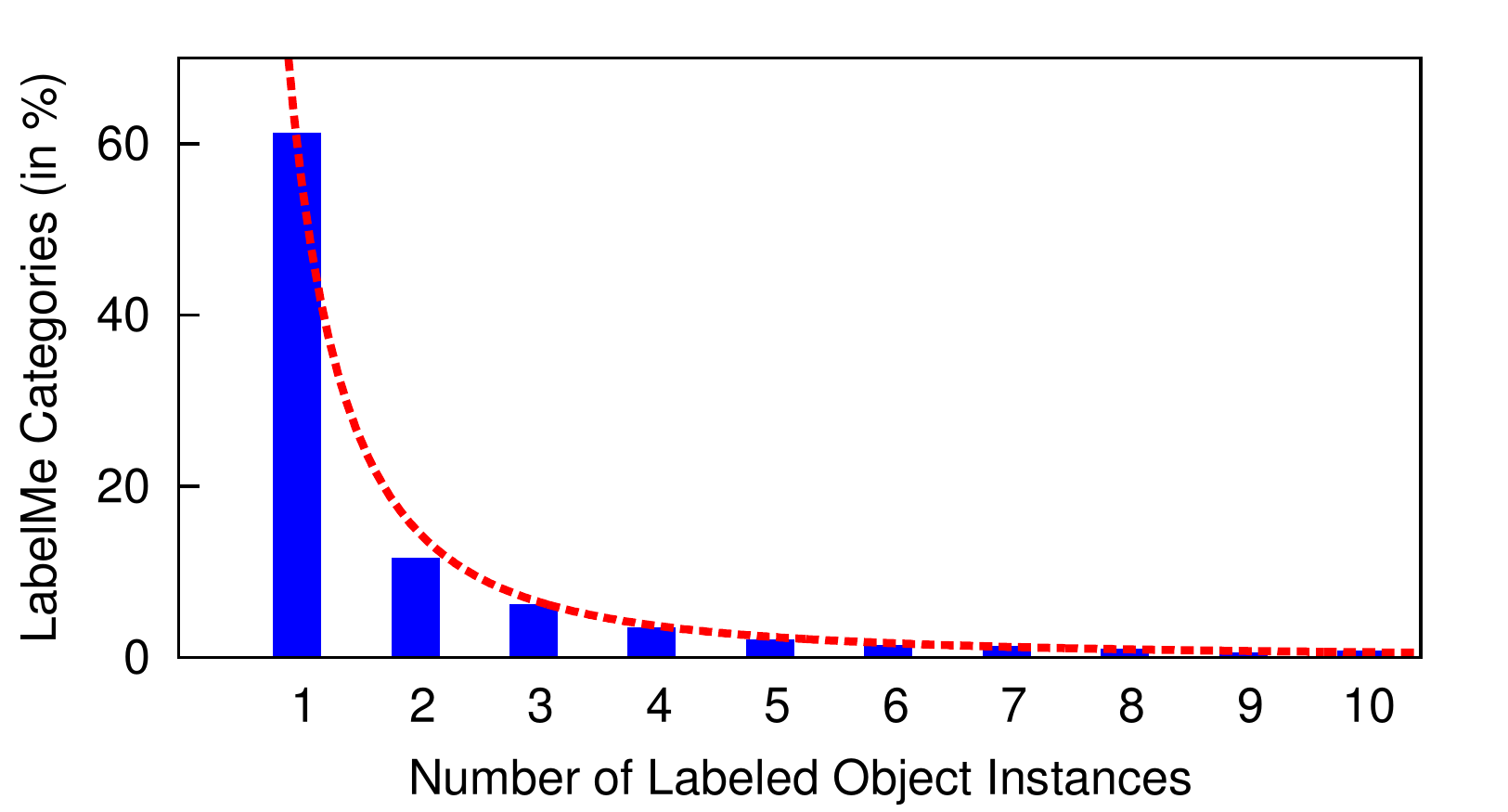}
		}
		\subfigure[]{
			\label{fig:zipflines}
			\includegraphics[angle=270,width=0.485\linewidth]{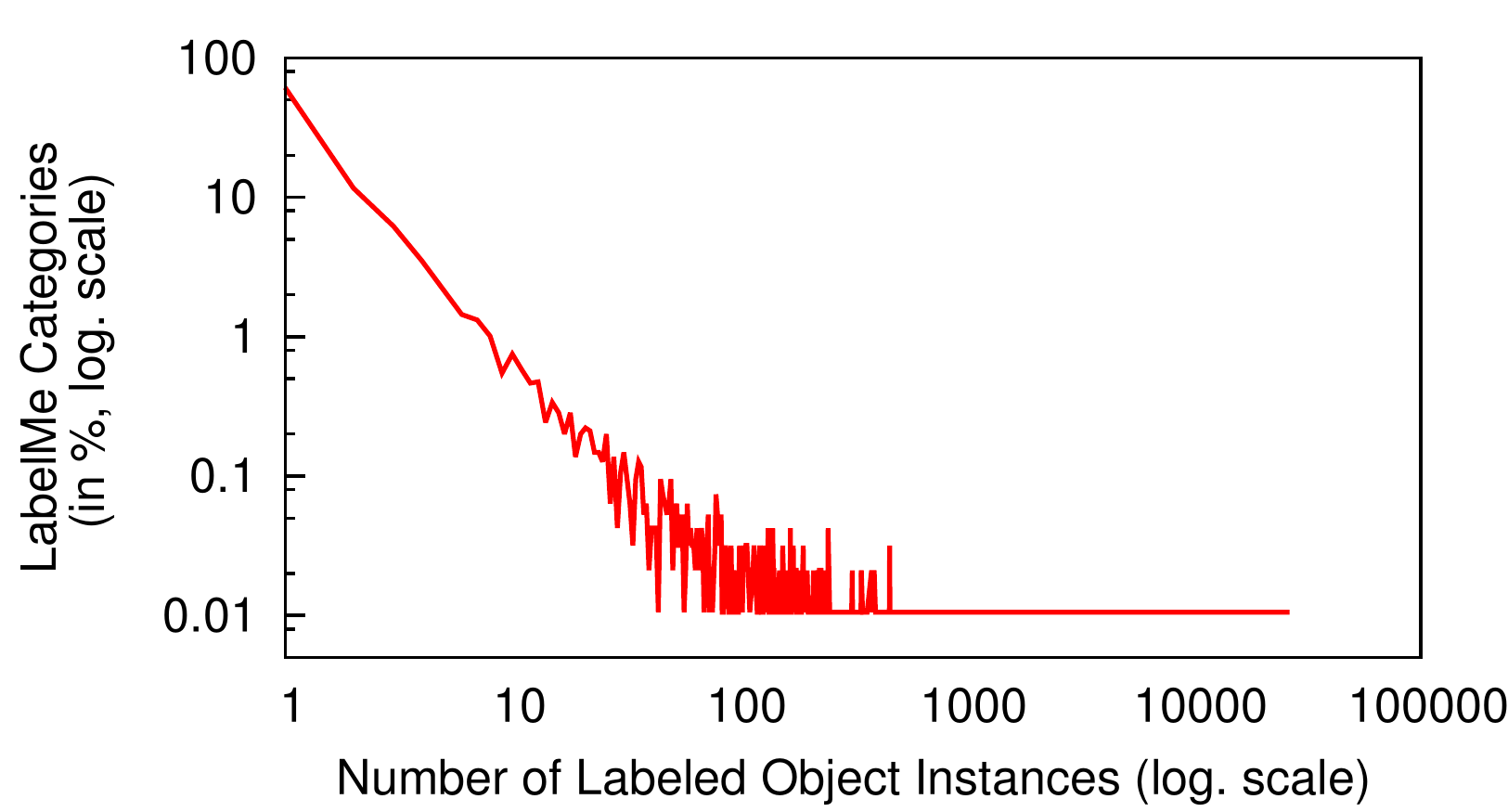}
		}
		\caption[Object category statistics and the law of Zipf.]{(a) Number of object categories in the LabelMe dataset for a specific number of labeled instances \emph{(inspired by \cite{Wang10:Cos})}; (b) Extended plot in logarithmic scale illustrating Zipf's law in the first part of the plot~\citep{Zipf49:HBa}. Similar statistics can be derived for \emphterm{ImageNet}~\citep{Deng09:INL} and
			\emphterm{TinyImages}~\citep{Torralba07:MTI}. The LabelMe database was obtained in April 2008.}
		\label{fig:zipf}
	\end{figure}

	The importance of efficient learning with few examples can be illustrated by analyzing current
	large-scale datasets for object recognition, such as \emphterm{LabelMe}~\citep{Russell08:LDW,Torralba10:Loi}.
	\figurename~\ref{fig:zipfbox} shows the relative number of object categories
	in LabelMe that possess a specific number of labeled instances. 
	A large	percentage (over $60$\%) of all categories only have one single labeled instance.
	Therefore, even in datasets which include an enormous number of images and annotations in total, 
	the lack of training data is a more common problem than one might expect.
	The plot also shows that the number of object categories with $k$ labeled instances follows
	an exponential function $\beta \cdot k^{-\alpha}$, which is additionally illustrated in the
	right log-log plot of \figurename~\ref{fig:zipf}. 
	This phenomenon is known as \newtermex{Zipf's law}{Zipf's law}~\citep{Zipf49:HBa} and can be found in language
	statistics and other scientific data.

	With current state-of-the-art approaches we are able to build robust object detectors for
	tasks with a large set of available training images, like detecting pedestrians or cars~\citep{Felzenszwalb08:DTM}. 
	However, if we want to extract a richer semantic representation of an image, such as trying to predict different
	visual attributes of a car (model type, specific identity, etc.), we are likely not
	able to rely on a sufficient number of images for each new category. Therefore, high-level 
	visual recognition approaches frequently suffer from weak training representations.

	\subsection{Industrial Applications}

	Problems related to a lack of learning examples are not restricted to visual object recognition tasks in 
	real-world scenes, but are	also prevalent in many industrial applications. 
	Collecting more training data is often expensive, time-consuming or practically impossible.
	In the following, we give three examples tasks in which such
	a problem arises.

	One important example is face identification~\citep{Tan06:FRS}, 
	where the goal is to estimate the identity of a person from a given image.
	For example, such a system has to be trained with images of each person
	being allowed to access a protected security area. 
	Obtaining hundreds of training images for each person is thus impractical, especially
	because the appearance should vary and include different illumination settings and clothing, which leads
	to a time-consuming procedure.
	Similar observations hold for writer or speaker verification~\citep{Yamada10:Ssi} and speech or handwritten text recognition~\citep{Bastien10:DSL}.
	
	Another interesting application scenario is the prediction of user preferences in shop
	systems.
	The goal is to estimate the probability that a client likes a new product, given some previous product
	selections and ratings. If a machine learning system quickly generalizes from a few given user ratings and achieves
	a high performance in suggesting good products to buy, it is more probable that the client
	will use this shop frequently. In this application area, solving the problem of learning with few training examples
	is simply a question of cost. The economical importance of the problem can be seen in
	the \newterm{Netflix prize}~\citep{Bennett07:np}, which promised one million dollars for a new
	algorithm which improves the rating accuracy of a DVD rental system. This competition has
	lead to a large amount of machine learning research related to \newterm{collaborative filtering},
	which is a special case of \newtermex{knowledge transfer}{transfer!knowledge|see{learning!transfer}} and is explained in more detail in Section~\ref{sec:knowledgetransfer}.
	
	A prominent and widely established field of application for machine learning and computer vision is
	\newtermabb{automatic visual inspection}{AVI}~\citep{Chin82:AVI}. To achieve a high quality of an industrial production, several work pieces
	have to be checked for errors or defects. Due to the required speed and the high cost of manual quality control,
	the need for automatic visual defect localization arises. Whereas images from non-defective data
	can be easily obtained in large numbers, training images for all kinds of defects are
	often impossible to collect. 
	A solution to solve this problem is to handle it as an \newtermex{outlier detection}{outlier detection|see{one-class classification}}
	or \newterm{one-class classification} problem. In this case, learning data only consists of non-defective
	examples and is used afterward to detect examples not belonging to the underlying distribution
	of the data. 

	\subsection{Challenges}
	\label{sec:challenges}
		What are the challenges and the problems of traditional machine learning methods in scenarios with few training examples?
		First of all, we have to clarify our notion of ``few''.
		Common to all traditional machine learning methods are their underlying assumptions,
		which were formulated by \cite{Niemann90:PAa}. The first postulate states:
		\begin{quote}
		\textbf{Postulate 1:} \textit{``In order to gather information about a task domain, a representative sample of patterns is available.'' \citep[\sectionname~1.3, p. 9]{Niemann90:PAa}}
		\end{quote}
		Therefore, a scenario with few training examples can be defined as a classification task that violates this assumption by
		having an insufficient or non-representative sample of patterns. Of course, this notion depends on the specific application
		and on the complexity of the task under consideration.
		\IncludeGraphicWithLongCaption{fig:imgcatdifficulty}{Images of two object categories}{Images of two object categories (\categoryName{fire truck} and \categoryName{okapi}) from the \emphterm{Caltech 256} database~\citep{Griffin07:Coc}}{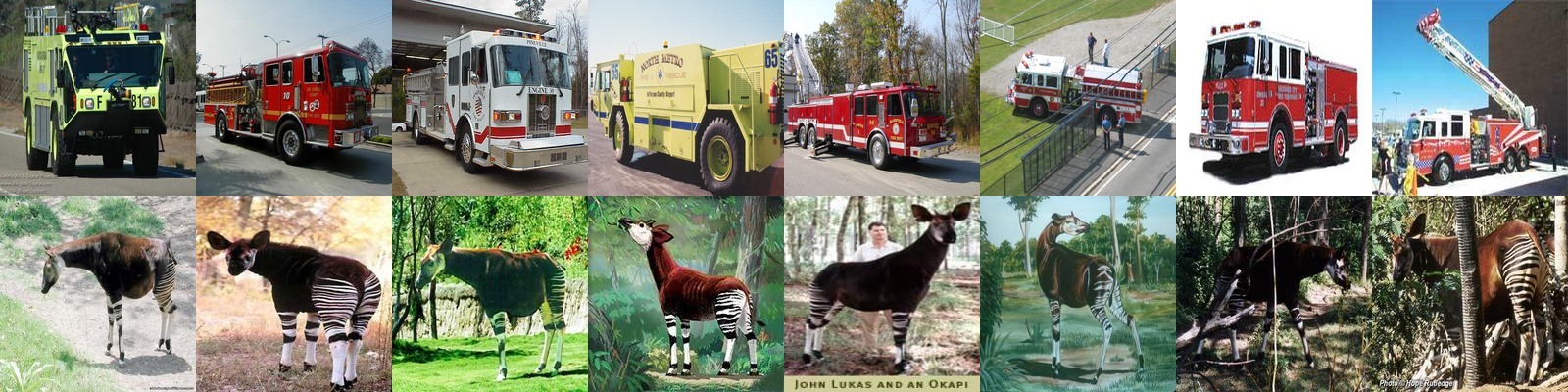}{0.9}

		One of the difficulties is the high variability in the data of high-level
		visual learning tasks. Some images from an object category database are given in \figurename~\ref{fig:imgcatdifficulty}. 
		A classification system has to cope with background clutter, different viewpoints, illumination changes and
		in general with a large diversity of the category (intra-class variance). 
		On the one hand, this can only be performed with a large amount of flexibility in the underlying 
		model of the category, such as using
		a large set of features extracted from the images and a complex classification model. 
		On the other hand, learning those models requires a large number of (representative) training examples.
		These conditions turn learning with few examples into
		a severe problem especially for high-level recognition tasks.

		The trade-off between a highly flexible model and the number of training examples required can be explained 
		quite intuitively for polynomial regression: Consider a set of $\numberOfExamples$ sample points of a one-dimensional $m$-order polynomial. 
		The order of the polynomial is a measure of the complexity of the function.
		For the noise-free case, we need $\numberOfExamples \geq m+1$ examples to get an exact and unique solution.
		In contrast, noisy input data requires a higher 
		number of examples to estimate a good approximation of the underlying polynomial.
		This direct dependency to the model complexity becomes more severe if we increase the input dimension $D$. 
		The number of coefficients that have to be estimated, and analogous the number of examples required, grows 
		polynomial in $\dimension$ according to ${ \dimension + m \choose m } = \landau( \dimension^m )$~\citep[Exercise 1.16]{Bishop07:PRM}.
		This immense increase in the amount of required training data is known in a broader as the \newterm{curse of dimensionality}.

		A deeper insight and an analysis for classification rather than regression tasks 
		is offered by the theoretical bounds derived in statistical learning theory for the error of a classification model or hypothesis 
		$\hypothesis$~\citep{Cucker02:mfo,Christianini00:ISV}.
		Assume that a learner selects a hypothesis from a possibly infinite set of hypotheses $\hypothesisSpace$ which achieves
		zero error on a given sampled training set of size $\numberOfExamples$. 
		The attribute ``sampled'' refers to the assumption that the training set is a sample from the underlying 
		joint distribution of examples and labels of the task. Due to this premise the following bound
		is not valid for one-class classification.
		A theorem proved by \citet[Corrolary 3.4]{Shawe-Taylor93:BSS}
		states that with probability of $1-\delta$ the following number of training examples
		is sufficient for achieving an error below $\epsilon$: 
		\begin{equation}
		\label{eq:minexamplesvc}
		\numberOfExamples \geq \frac{1}{\epsilon (1-\sqrt{\epsilon})} \left(2 \ln \left( \frac{6}{\epsilon} \right) \dim(\hypothesisSpace) + \ln \left( \frac{2}{\delta} \right) \right) \enspace.
		\end{equation}
		The term $\dim(\hypothesisSpace)$ denotes the \newterm{VC dimension}\footnote{VC is an abbreviation for Vapnik–Chervonenkis.} of $\hypothesisSpace$ and can be regarded as a complexity measure of
		the set of available models or hypotheses. For example, the class of all $D$-dimensional linear classifiers including a bias term
		has a VC dimension of $D+1$~\citep[\sectionname~4.11, Example 2]{Vapnik00:NSL}.
		Let us now take a closer look on the bound in \equationname~\eqref{eq:minexamplesvc}. If we fix the maximum error $\epsilon$
		and choose an appropriate small value for $\delta$, we can see that the sufficient number of training examples depends linearly 
		on the VC dimension $\dim(\mathbb{H})$.
		This directly corresponds to our previous example of polynomial regression, because the VC dimension of $D$-variate polynomials
		of up to order $m$ is exactly ${D+m \choose m}$ \citep{Ben-David98:LvI}.

	\subsection{The Importance of Prior Knowledge}
		Nearly all machine learning algorithms can be formulated as optimization problems, whether in a direct way, such as done by \newtermabb{support vector machines}{SVM}~\citep{Christianini00:ISV}, 
		or in an indirect manner like \newterm{boosting}~\citep{Friedman00:ALR} approaches.
		From this point of view, we can say that learning with few examples 
		inherently tries to solve an ill-posed optimization problem.
		Therefore, it is not possible to find a suitable well-defined solution without incorporating additional (prior) information.
		The role of prior information is to (indirectly) reduce the set of possible hypotheses. For example, if we know in advance that for a classification
		task only the color of an object is important, \eg if we want to detect expired meat, only a small number of features 
		have to be computed and a lower dimensional linear classifier can be used. In this situation the VC dimension is reduced,
		which results in a lower bound for the sufficient number of training examples (\cf \equationname~\eqref{eq:minexamplesvc}).
		
		Introducing common prior knowledge into the machine learning part of a visual recognition system
		is often done by regularization techniques that penalize non-smooth solutions~\citep{Scholkopf01:LwK} or the
		``complexity'' of the classification model. 
		Examples of such techniques are $L_2$-regularization, also known as \newtermex{Tikhonov regularization}{regularization!Tikhonov}~\citep{Vapnik00:NSL}, or $L_1$-regularization,
		which is mostly related to methods trying to find a sparse solution~\citep{Seeger08:BIa}.

		Other possibilities to incorporate prior knowledge include \newtermex{semi-supervised learning}{learning!semi-supervised} and \newtermex{transductive learning}{learning!transductive}, which utilize large
		sets of unlabeled data to support the learning process~\citep{Fergus09:SLi}. Unlabeled data can help to estimate
		the underlying data manifold and, therefore, are able to reduce the number of model parameters. 
		However, the use of unlabeled data is not studied in this thesis.

\section{Knowledge Transfer and Transfer Learning}
\label{sec:knowledgetransfer}
	
	Machine learning tasks related to computer vision always require a large amount of prior knowledge. 
	For a specific task, we indirectly incorporate prior knowledge into the classification system by choosing
	image preprocessing steps, feature types, feature reduction techniques, or
	the classifier model. This choice is mostly based on prior
	knowledge manually obtained by a software developer from previous experiences on similar
	visual classification tasks. For instance, when developing an automatic license plate reader, ideas
	can be borrowed from optical text recognition or traffic sign detection.
	Increasing expert prior knowledge decreases the number of training examples needed by the classification system
	to perform a learning task with a sufficient error rate. 
	However, this requires a large manual effort.
	
	The goal of some techniques presented in this work is to perform transfer of prior knowledge
	from previously learned tasks to a new classification task in an automated
	manner, which is known as \newtermex{transfer learning}{learning!transfer} and which is 
	a special case of \newtermex{knowledge transfer}{transfer!knowledge}. 
	The advantage compared to traditional machine learning methods, or \newtermex{independent learning}{learning!independent}, is that 
	we do not have to build new classification systems from the scratch or by large manual effort.
	Previously known tasks used to obtain prior knowledge are referred to as \newtermex{support tasks}{support task} and a new classification
	task only equipped with few training examples is called \newterm{target task}.
	In the following, we concentrate on \newtermex{inductive transfer learning}{learning!inductive transfer}~\citep{Pan10:STL}, which assumes
	that we have labeled data for the target and the support tasks.
	Especially interesting are situations where a large number of training examples for the support tasks exists and
	prior knowledge can be robustly estimated.
	Other terms for transfer learning are \newtermex{learning to learn}{learning!to learn}~\citep{Thrun97:LTL}, \newtermex{lifelong learning}{learning!lifelong} and
	\newtermex{interclass transfer}{transfer!interclass}~\citep{Fei-Fei06:OLO}.

	\IncludeGraphicWithLongCaption{fig:simpleillu}{The basic idea of knowledge transfer for visual object recognition}
		{The basic idea of transfer learning for visual object recognition: a lot of visual
		 categories share visual properties which can be exploited to learn a new object category from few training examples.}
		{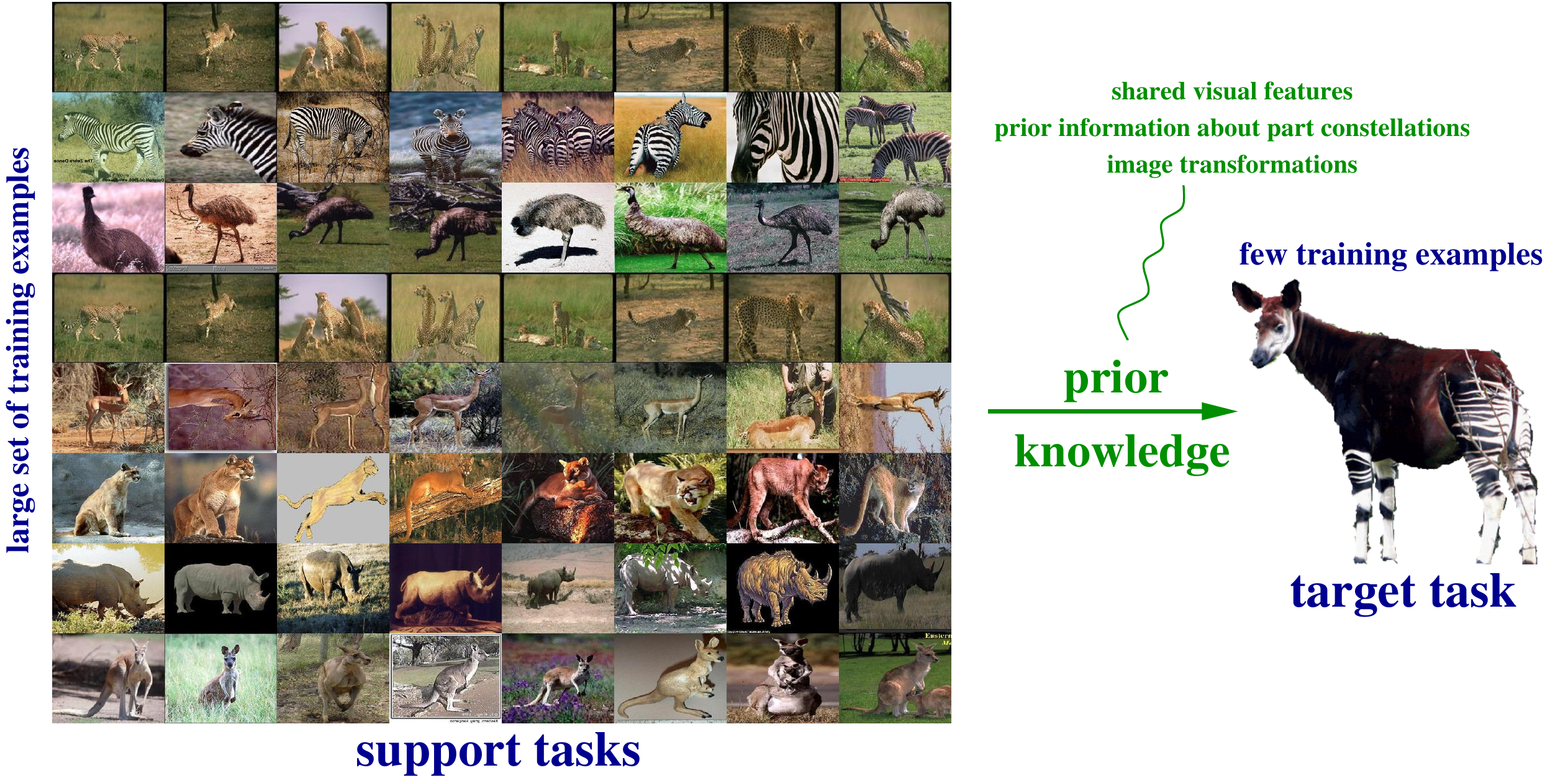}{0.8}
	The concept of transfer learning is also one of the main principles that explain the development of the human
	perception and recognition system~\citep{Brown88:PCC}. For example, it is much easier to learn Spanish if
	we are already able to understand French or Italian. The knowledge transfer concept is known in the language domain
	as language transfer or linguistic interference~\citep{Odlin89:LtC}.
	We already mentioned that a child quickly learns new visual object categories in an incremental manner without
	using many learning examples.
	\figurename~\ref{fig:simpleillu} shows some images of a transfer learning scenario for visual object recognition. 
	Generalization from a single example of the new animal category, on the right hand side of \figurename~\ref{fig:simpleillu},
	is possible due to a large set of available memories (images) of related animal categories.
	These categories often share visual properties, such as typical object part constellations (head, body and four legs)
	or similar fur texture.

	Developing transfer learning techniques and ideas requires answering four different questions: \textit{``What, how, from where and when to transfer?''}.
	First of all, the type of knowledge which will be transferred from support tasks to a new target task has to be defined, \eg
	information about common suitable features. Detailed examples are listed in a paragraph of \sectionname~\ref{sec:whathow}. 
	The transfer technique applied to incorporate prior knowledge into the learning process of the target task strictly
	depends on this definition but is not determined by it. For example, the relevance of features for a classification task can
	be transferred using generalized linear models~\citep{Lee07:LML} or random decision forests.
	Prior knowledge is only helpful for a target task if the support tasks are somehow related or similar.
	In some applications not all available previous tasks can be used as support tasks, because
	they would violate this assumption. 
	Giving an answer to the question \textit{``From where to transfer?''} means that the learning algorithm
	has to select suitable support tasks from a large set of tasks. These learning situations are referred to
	as learning in \newterm{heterogeneous environments}.
	Of course, we expect that additional information incorporated by transfer learning always improves 
	the recognition performance on a new target task, because it is the working hypothesis
	of transfer learning in general. However, in machine learning there is no guarantee at all that
	the model learned from a training set is also appropriate for all unseen examples of the task,
	\ie there are no deterministic warranties concerning the generalization ability of a learner 
	\footnote{Note that even the bound in \equationname~\eqref{eq:minexamplesvc} only holds with probability $1-\delta$.}.
	Therefore, knowledge transfer can fail and lead to worse performance compared to independent learning.
	This event is known as \newtermex{negative transfer}{transfer!negative} and happens in everyday life. For example, if we use ``false friends'' when
	learning a new language, \eg German speakers are sometimes confused about ``getting a gift'' because the word ``Gift'' is the
	German word for poison, which is likely not a thing you are happy to get. 
	Situations in which negative transfer might occur are difficult to detect. 

	\begin{figure}[tb]
		\centering
		\subfigure[Independent Learning]{
			\includegraphics[width=0.45\linewidth]{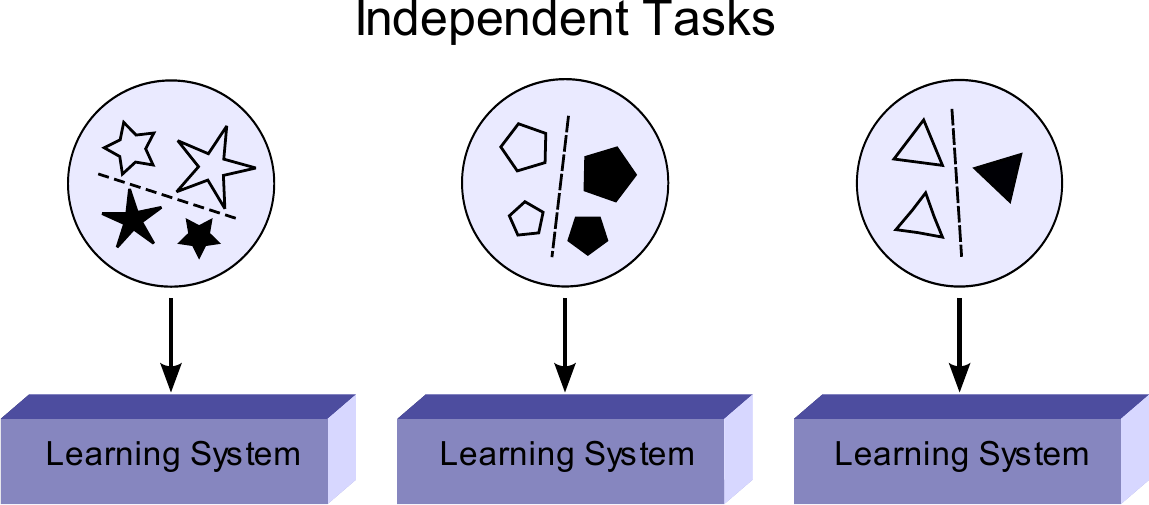}
		}\hfill
		\subfigure[Transfer Learning]{
			\includegraphics[width=0.45\linewidth]{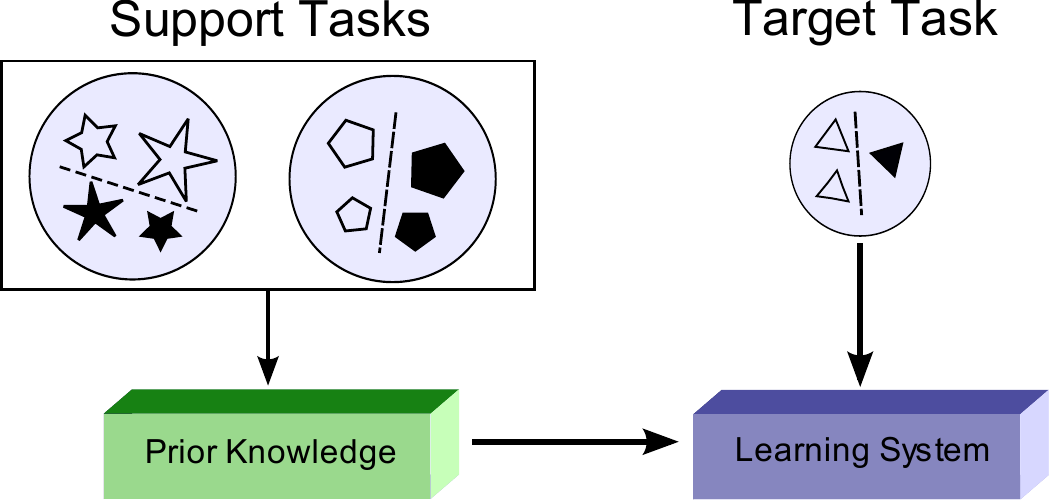}
		}\hfill
		\subfigure[Multitask Learning]{
			\includegraphics[width=0.45\linewidth]{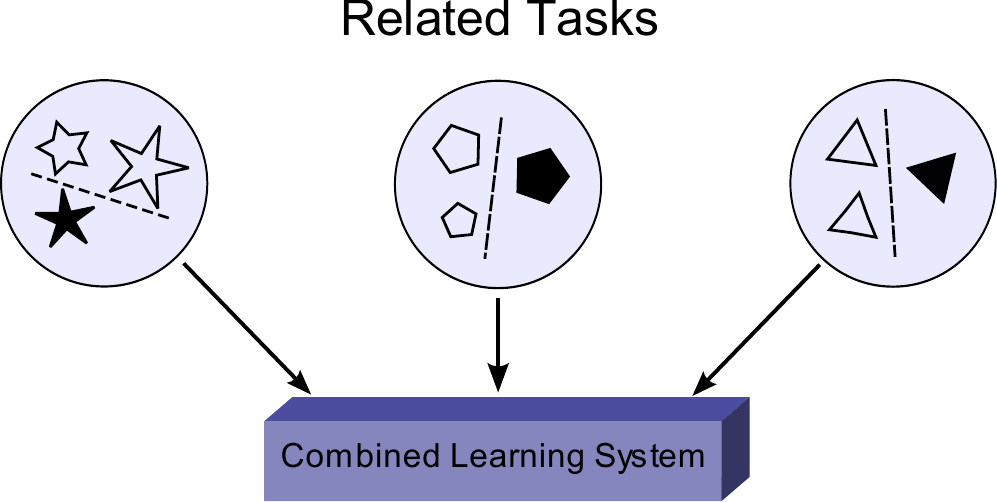}
		}\hfill
		\subfigure[Multi-class Transfer Learning]{
			\includegraphics[width=0.45\linewidth]{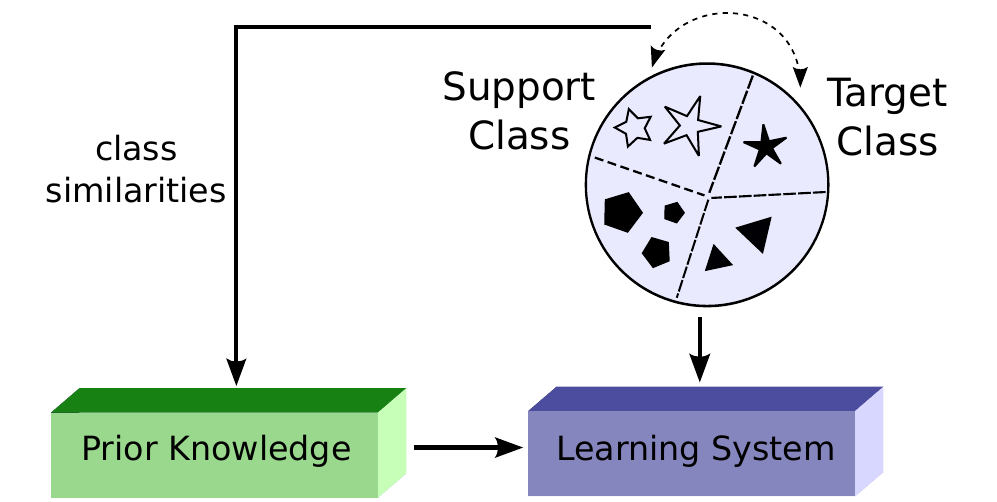}
		}
		\caption[Main idea of transfer learning]{The concept of (a) traditional machine learning with independent tasks, (b) transfer learning, 
			(c) multitask learning and (d) multi-class transfer learning \textit{(inspired by \cite{Pan10:STL})}.}
		\label{fig:concepts}
	\end{figure}
	Besides transfer learning, another type of knowledge transfer is \newtermex{multitask learning}{learning!multitask} which learns different classification tasks
	jointly\footnote{\citet{Xue07:MLC} use the term symmetric multitask learning to refer to jointly learning tasks and asymmetric multitask learning refers to our use
	of the term transfer learning.}. Combined estimation of model parameters can be very helpful, especially if a set of tasks are given, with each
	having only a small number of training examples. In contrast to transfer learning there is no prior knowledge obtained in advance, but
	the model parameters of each task are coupled together. For example, given a set of classification tasks, relevant features can be estimated
	jointly and all classifiers are learned independently with the reduced set of features.
	A possible application is collaborative filtering as mentioned in \sectionname~\ref{sec:motivation}.
	In the following thesis, we stick to transfer learning but borrow some ideas from multitask learning approaches.

	\figurename~\ref{fig:concepts} summarizes the conceptual difference between independent learning, transfer learning, and multitask learning.
	Furthermore, we illustrate the principle of \newtermex{multi-class transfer learning}{learning!multi-class transfer}. In contrast
	to all other approaches, transfer within a single multi-class task is considered rather than between several (binary) classification tasks.
	To emphasize this fact, we use the term target class rather than target task.
	The main difficulty is that the target class has to be distinguished from the support class, even though information was transferred and exploits
	their similarity.

	It should be noted that another area of knowledge transfer is \newtermex{transductive transfer learning}{learning!transductive transfer}, which concentrates on transferring knowledge
	from one application domain to another. For example, the goal is to recognize objects in low quality webcam images with the support
	of labeled data from photos made by digital cameras. Related terms are \newterm{sample-selection bias}, \newterm{covariate shift}, 
	and \newterm{domain adaptation}~\citep{Pan10:STL}. 

	\section{Literature Overview: Transfer Learning}
	\label{sec:previouswork}
      
	There is a large body of literature trying to handle the
	problem of learning with few examples. A lot of work concentrates on
	new feature extraction methods, or classifiers, which show superior
	performance to traditional methods especially for few training examples~\citep{Levi04:LOD}.
	The few examples problem was also tackled by introducing manual prior knowledge
	of the application domain, such as augmenting the training data by
	applying artificial transformations~\citep{Duda00:PCE,Bayoudh07:LCV} also
	known as \newterm{data manufacturing}.

	In the following, we concentrate on methods related to
	the principle of transfer learning and multitask learning as introduced in the previous section.
	Other similar surveys and literature reviews can be found in the work
	of \citet{Fei-Fei06:KTI} from a computer vision perspective and
	the journal paper of \citet{Pan10:STL}, which gives
	a comprehensive overview of the work done in the machine learning community.
	There is also a textbook by~\cite{Brazdil09:MA} covering the broader area
	of \newtermex{meta-learning}{learning!meta}, and the edited book of~\cite{Thrun97:LTL}, about
	the early developments of learning to learn.

	\subsection{What and how to transfer?}
	\label{sec:whathow}

		The type of knowledge which is transferred from support tasks to a target task
		is often directly coupled with the method used to incorporate this additional
		information. Therefore, we give a combined literature review on answers to 
		both questions.

		\paragraph{Learning Transformations}
		\label{sec:congealing}
		One of the most intuitive types of knowledge which can be transferred between
		categories is application-specific transformations or distortions.
		While in data manufacturing methods these transformations have to
		be defined using manual supervision, transfer learning methods
		learn this information from support tasks.

		For example, a face recognition application can significantly
		benefit from transformations transferred from other
		persons using optical flow techniques \citep{Beymer95:FRO}.
		Estimating latent transformations and distortions of images
		(e.g. illumination changes, rotations, translations)
		within a category is 
		proposed by \citet{Miller00:LOE} and \citet{Learned-Miller06:DDI} using an 
		optimization approach.
		Their approach called \newterm{Congealing} tries to minimize the
		joint entropy of gray-value histograms in each pixel with a greedy
		strategy. The obtained transformations can be directly applied
		to the images of a target task and used in a nearest neighbor approach
		for text recognition.
		Restricting and regularizing the complexity of the class of transformations during
		estimation is important for a good generalization, because it
		additionally introduces generic prior knowledge.
		The original Congealing approach proposed a heuristic normalization step
		directly applied during optimization. An extension without explicit normalization,
		and a study of different complexity measures, can be found in the
		work of \citet{Vedaldi08:JAL} and \citet{Vedaldi07:CDA}.
		\citet{Huang07:UJA} use Congealing to align
		images of cars or faces with local features.

		\paragraph{Shared Kernel or Similarity Measure}
	
		How we compare objects and images strictly depends on the current task.
		A distance metric or a more general similarity measure can be an important
		part of a classification system, e.g. in nearest neighbor approaches.
		The term \newterm{kernel} is a related concept which also measures
		the similarity between input patterns.
		Hence, a distance metric or a kernel function is an important piece of
		prior knowledge which can be transferred to new tasks.
		
		The early work of \cite{Thrun96:LTA} used neural network techniques
		to estimate a similarity measure for a specific task.  
		In general, the idea of estimating an appropriate metric
		from data is a research topic on its own called \newterm{metric learning}~\citep{Yang06:DML}. 
		A common idea is to find a metric which minimizes distances
		between examples of a single category (intra-class) and maximizes distances
		between different categories (inter-class).
		Applying a similarity measure to another task is straightforward when
		using a nearest neighbor classifier.
		\cite{Fink04:OCS} used the metric learning algorithm of \cite{Shalev-Shwartz04:OBL},
		which allows online learning and estimates the correlation matrix of a 
		Mahalanobis distance.
	
		Metric learning for visual identification tasks is presented by~\cite{Ferencz08:LLI}.
		They show how to find discriminative local features which can be used 
		to compare different instances of an object category, \eg distinguishing between specific instances of a car. 
		In this work, metric learning is based 
		on learning a binary classifier which estimates the probability that both
		images belong to the same object instance. 
		The obtained similarity measure can be used for visual identification with only
		one single training example for each instance.
		Another application of metric learning is domain adaptation as explained in 
		\sectionname~\ref{sec:knowledgetransfer}. 
		The paper of \citet{Saenko10:AVC} presents a new database for testing domain
		adaptation methods and also gives results of a metric learning algorithm.
		In contrast, \citet{Jain11:ODA} propose to perform domain adaptation by applying Gaussian process regression 
		on scores of examples near the decision boundary. 
		
		\paragraph{Shared Features}
		
		Visual appearance can be described in terms of features such as color, shape and 
		local parts. Thus, it is
		natural to transfer information about the relevance of features for a given task.
		This idea can be generalized to shared base classifiers which allow modeling
		feature relevance. 
		\cite{Fink06:OML} study combining perceptron-like classifiers of the support and
		target task. 
		Due to the decomposition into multiple base classifiers (weak learners), ensemble methods and especially
		Boosting approaches~\citep{Freund97:DGO} are suitable for this type of knowledge transfer.
		\citet{Levi04:LSN} extend the standard Boosting framework by integrating task-level
		error terms. Weak learners which achieve a small error on all tasks should be preferred to
		very specific ones.
		A similar concept is used in the work of \citet{Torralba07:SVF}, who propose learning
		multiple binary classifiers jointly with a Boosting extension called \emphterm{JointBoost}. 
		The algorithm tries to find weak learners shared by multiple categories. This also
		leads to a reduction of the computation time needed to localize an object with a sliding-window
		approach. Experiments of \citet{Torralba07:SVF} show that 
		the number of feature evaluations grows logarithmic in the number of categories,
		which is an important benefit compared to independent learning.
		An extension of this approach to kernel learning can be found in~\citet{Jiang07:KSW}.
		\citet{Salakhutdinov11:LtS} exploits category hierarchies and performs feature sharing 
		by combining hyperplane parameters.
		 
		\paragraph{Shared Latent Space}
		
		Finding discriminative and meaningful features is a very difficult task.
		Therefore, the assumption of shared features between tasks is often not valid
		for empirically selected features used in the application.
		\citet{Quattoni07:LVR} propose assuming an underlying latent feature space which is common to all tasks.
		They use the method of \citet{Ando05:FLP} to estimate a feature transformation from support
		tasks derived from caption texts of news images.
		To estimate relevant features, the subsequent work~\citep{Quattoni08:TlI} propose an
		eigenvalue analysis and the use of unlabeled data.

		Latent feature spaces can be modeled in a Bayesian framework using Gaussian processes,
		which leads to the so called \newtermabb{Gaussian process latent variable model}{GP-LVM}~\citep{Lawrence05:Pnp}.
		Incorporating the idea of a shared latent space into the \mbox{GP-LVM} framework
		allows using various kinds of noise models and kernels~\citep{Urtasun08:TNR}.

		\paragraph{Constellation Model and Hierarchical Bayesian Learning}
		
		An approach for knowledge transfer between visual object categories was presented by \citet{Fei-Fei06:OLO}.
		Their method is inspired by the fact that a lot of categories share common object parts which are
		often also in the same relative position. Based on a generative constellation model~\citep{Fergus03:OCR} they
		propose using maximum a posteriori estimation to obtain model parameters for a new target category.
		The prior distribution of the parameters corresponding to relative part positions
		and part appearance is learned from support tasks.
		The underlying idea can also be applied to a shape based approach~\citep{Stark09:SOC}.
		
		A prior on parameters shared between tasks is an instance of hierarchical Bayesian learning.
		\cite{Raina06:TLC} used this concept to transfer covariance estimates of parts of the feature
		set.

		\paragraph{Joint Regularization}
		\index{regularization!joint}
		A lot of machine learning algorithms such as SVM are not directly formulated in a probabilistic
		manner but as optimization problems. These problems often include
		regularization terms connected to the complexity of the parameter, which
		would correspond to a prior distribution in a Bayesian setting.
		The equivalent to hierarchical Bayesian learning as described in the last paragraph
		is a joint regularization term shared between tasks.

		\cite{Amit07:USS} propose using \newtermex{trace norm regularization}{regularization!trace norm} of
		the weight matrix in a multi-class SVM approach. They show that this regularization
		is related to the assumption of a shared feature transformation and task-specific weight vectors
		with independent regularization terms. Instead of transferring knowledge between
		different classification tasks this work concentrates on transfer learning in a multi-class
		setting, \ie multi-class transfer.

		Sharing a low dimensional data representation for multitask learning is the idea of \cite{Argyriou06:MTF}.
		The proposed optimization problem learns a feature transformation and a weight vector jointly and
		additionally favors sparse solutions by utilizing an $L_{1,2}$ regularization.
		Multitask learning with kernel machines was first studied by \cite{Evgeniou05:LMT}.
		Their idea is to reduce the multitask problem to a single task setting by defining a combined kernel function 
		or \newtermex{multitask kernel}{kernel!multitask} and a new regularizer.

		\paragraph{Shared Prior on Latent Functions}

		The framework of Gaussian processes allows modeling a prior distribution of an underlying
		latent function for	each classifier~\citep{Rasmussen05:GP} using a kernel function. 

		If we want to learn a set of classifiers jointly in a multitask setting, an appropriate
		assumption is that all corresponding latent functions are sampled from the same prior
		distribution. \cite{Lawrence04:EoI} suggest learning the hyperparameters of the kernel
		function jointly by maximizing the marginal likelihood.
		This idea was also applied to image categorization tasks~\citep{Kapoor09:GPO}.
		A more powerful way of performing transfer learning is studied with multitask kernels originally introduced
		by \cite{Evgeniou05:LMT}.
		\cite{Bonilla07:KMT} use a parameterized multitask kernel that is the product of
		a base kernel comparing input features and a task kernel modeling the similarity between
		tasks and using meta or \newterm{task-specific features}.
		Task similarities can also be learned without additional meta features by estimating a non-parametric 
		version of the task kernel matrix~\citep{Bonilla08:MTG}.
		A theoretical study of the generalization bounds induced by this framework can be found in \cite{Chai08:MGP}.
		\cite{Schwaighofer05:LGp} propose an algorithm and model to learn the fully non-parametric form of the multitask kernel 
		in a hierarchical Bayesian framework.
		
		The \newtermabb{semi-parametric latent factor model}{SLFM} of \cite{Teh05:SLF} is directly related to
		a multitask kernel. The latent function for each task is modeled as
		a linear combination of a smaller set of underlying latent functions.
		Therefore, the full covariance matrix has a smaller rank, which directly corresponds to
		the rank assumption of other transfer learning ideas~\citep{Amit07:USS}.
		A more general framework which allows modeling
		arbitrary dependencies between examples and tasks using a graph-theoretic notation is
		presented by~\cite{Yu08:GPM}.

		\paragraph{Semantic Attributes and Similarities}
		Transfer learning with very few examples of the target task
		can be difficult due to the lack of data to estimate task relations and similarities 
		correctly. Especially if no training data (neither labeled nor unlabeled) is
		available, other data sources have to be used to perform transfer learning.
		This scenario is known as \newtermex{zero-shot learning}{learning!zero-shot} and uses the concept of
		\newtermex{learning with attributes}{learning!with attributes}, an area which 
		received much attention in recent years. 
		The term attribute refers to category-specific features.

		\cite{Lampert09:LDU} use a large database of human-labeled abstract attributes of animal classes
		(e.g. brown, stripes, water, eats fish). One idea is to train several attribute classifiers 
		and use their output as new meta features. 
		This representation allows recognizing new categories without real training images
		only by comparison with the attribute description of the category.
		The knowledge transferred from support tasks is the powerful discriminative
		attribute representation which was learned with all training data.
		A similar idea is presented by \cite{Larochelle08:Zlo} for zero-shot learning
		based on task-specific features.
		A theoretical investigation of zero-shot learning with an attribute representation
		is given by \cite{Palatucci09:ZLw} and concentrates on analysis with the concept
		of \newtermabb{probable approximate correctness}{PAC}~\citep{Vapnik00:NSL}.
		
		Instead of relying on human-labeled attributes,
		internet sources can be used to mine attributes and relations.
		The papers of \cite{Rohrbach10:CLS,Rohrbach10:WHW} compare different kinds
		of linguistic sources, such as WordNet~\citep{Pedersen04:WNS}, Google search, Yahoo and Flickr.
		A large-scale evaluation of their approach can be found in \citet{Rohrbach11:EKT}.
		Attribute based recognition can help to generalize 
		to new tasks or categories \citep{Farhadi09:DOA} which is otherwise difficult
		using a training set only equipped with ordinary category labels.
		Attributes can also help to boost the performance of
		object detection rather than image categorization as shown in \citep{Farhadi10:ARC}.
		Their transfer learning approach heavily relies on model sharing of object parts
		between categories.
		
		\cite{Lampert10:WMC} use a generalization of maximum covariance analysis to 
		find a shared latent subspace of different data modalities. This can also be applied
		to transfer learning with attributes by regarding the attribute representation
		as a second modality.
		Beyond zero-shot learning semantic similarities are also
		used to guide regularization~\citep{Wang10:Cos}.
	  
		\paragraph{Context Information}
		
		Up to now, we only covered transfer learning in which knowledge
		is used from visually similar object categories or tasks.
		However, dissimilar categories can also provide useful information if
		they can be used to derive contextual information. For example
		it is likely to find a keyboard next to a computer monitor, which can be
		a valuable information for an object detector.
		Methods using contextual information always exploit dependencies between
		categories and tasks and are therefore a special case for knowledge transfer approaches.

		\cite{Fink03:MBC} propose training a set of object detectors
		simultaneously with an extended version of the \emphterm{AdaBoost} algorithm~\citep{Viola04:RRO}
		and can be regarded as an instance of multitask learning.
		In each round of the boosting algorithm the map of detection scores 
		is updated and used as an additional feature in subsequent rounds.
		A similar idea is presented by \cite{Shotton08:STF} for \newterm{semantic segmentation},
		which labels each pixel of the image as one of the learned categories. 
		The work of \cite{Hoiem05:GCS} pursues the same line of research, but
		clearly separates the support and target tasks.
		In a first step geometric properties of image areas are estimated.
		The resulting labeling into planar, non-planar, and porous objects, as well as ground and sky areas
		can be used to further assist local detectors as high-level features.
		Contextual relationships between different categories can also be modeled directly
		with a \newtermabb{conditional Markov random field}{CRF} as done by \cite{Rabinovich07:OiC} and \cite{Galleguillos08:WSO}
		on a region-based level for semantic segmentation. 
		
	\subsection{Heterogeneous Transfer: From Where to Transfer?}

		Automatically selecting appropriate support tasks from a large set
		is a difficult sub-task of transfer learning. Therefore, most of the previous work presented in this
		thesis so far assumes that support tasks are given in advance.
		An exception is the early work of \cite{Thrun96:DSI}, which proposes the \newterm{task clustering} algorithm. 
		Similarities between two tasks are estimated by testing the classifier learned on one task 
		using data from the other task. Afterward, clustering can be performed with the resulting task similarity matrix.

		\cite{Mierswa05:ECB} select relevant features for a target task by comparing 
		the weights of the SVM hyperplane with each of the available tasks. Therefore, the algorithm
		selects a similar but more robustly estimated feature representation.
		The work of \cite{Kaski07:LRT} performs transfer learning with logistic regression classifiers
		and models the likelihood of each task as a mixture of the target task likelihood and a likelihood
		term which is independent of all other tasks. Due to the task-dependent weight, the algorithm can
		adapt to heterogeneous environments.
		In general, selecting support tasks is a model selection problem, therefore, 
		techniques like leave-one-out are used~\citep{Tommasi09:myk,Tommasi10:SiN}.
		Heterogeneous tasks can also be handled within the regularization framework
		of \cite{Argyriou06:MTF}
		by directly optimizing a clustering of the tasks~\citep{Argyriou08:ATL}.

\section{Summary}

	We gave a summary of current work done in the area of visual transfer learning. Although there is a huge number of papers dealing with the problem of transferring information between tasks (or domains), many of the methods share the same assumptions and underlying basic ideas.


\end{document}